\documentclass{article}

\PassOptionsToPackage{numbers}{natbib}
\usepackage[final]{airov25}

\usepackage[utf8]{inputenc} 
\usepackage[T1]{fontenc}    
\usepackage{booktabs}       
\usepackage{amsfonts}       
\usepackage{nicefrac}       
\usepackage{microtype}      
\usepackage{xcolor}         

\usepackage{multirow}
\usepackage{graphicx}
 \usepackage{amsmath}
 \usepackage[utf8]{inputenc}
 \usepackage{subcaption}

\title{Assessing the impact of Binarization for Writer
Identification in Greek Papyrus}

\author{%
  Dominic Akt \\
  TU Wien \\
  Vienna, Austria\\
  \And
    Marco Peer \\
   HEIA-FR\\ 
    Fribourg, Switzerland\\
  \And
  Florian Kleber \\
  TU Wien \\
  Vienna, Austria\\
  }

\begin{document}

\maketitle

\begin{abstract}
  This paper tackles the task of writer identification for Greek papyri.
  A common preprocessing step in writer identification pipelines is image binarization, which prevents the model from learning background features.
  This is challenging in historical documents, in our case Greek papyri, as background is often non-uniform, fragmented, and discolored with visible fiber structures. We compare traditional binarization methods to state-of-the-art Deep Learning (DL) models, evaluating the impact of binarization quality on subsequent writer identification performance. DL models are trained with and without a custom data augmentation technique, as well as  different model selection criteria are applied. The performance of these binarization methods, is then systematically evaluated on the DIBCO 2019 dataset. The impact of binarization on writer identification is subsequently evaluated using a state-of-the-art approach for writer identification.
  The results of this analysis highlight the influence of data augmentation for DL methods. Furthermore, findings indicate a strong correlation between binarization effectiveness on papyri documents of DIBCO 2019 and downstream writer identification performance.
\end{abstract}

\section{Introduction} \label{introduction}
Writer retrieval and writer classification are tasks in the field of
document analysis, with applications ranging from historical research to
forensics \citep{kairacters}. Writer retrieval aims to identify documents
written by the same writer, based on handwriting similarities, by
returning a ranked list of potential matches. In contrast, writer
classification seeks to assign a given document image to one author from
a predefined set of known writers by returning a single author label.
Although writer identification and writer classification are sometimes used
interchangeably in the literature, this work defines ``writer
identification'' as an umbrella term encompassing both writer retrieval
and writer classification tasks.

Deep Learning (DL) approaches, particularly unsupervised DL methods, have demonstrated comparable or even superior performance to ``classical'' supervised techniques in paleographic tasks \citep{jimaging6090089}. A fundamental preprocessing step for most
DL-based writer identification approaches is image binarization
\citep{christlein2022writer}, which converts color or grayscale document
images into binary images, separating text (foreground) from
background. Binarization is essential because it prevents models from
learning background features, allowing the model to focus exclusively on
the characteristics of the author's handwriting.

The challenges and potential benefits of automated writer identification
become particularly evident when dealing with ancient papyri.
Paleographic experts face the challenge of analyzing large collections
of ancient papyri, many of which lack author attribution or dates
\citep{grk_papyri}. Specifically, writer retrieval can assist experts by
automatically grouping unattributed fragments that may belong together,
enabling the reconstruction of larger texts or identifying the
output of a single scribe. Writer classification, on the other hand, can
help experts attribute newly found or analyzed fragments to known
writers. DL methods are the current state of the art for writer identification \citep{DLwriterretrieval}.

While binarization is a well-studied problem for machine printed text or
handwriting on paper, applying it to historical documents such as papyri
presents unique challenges. Papyri can suffer from severe degradation as
some of the earliest Greek papyri found date back to 400 BCE
\citep{turner2015greek}. They exhibit degradations such as faded text,
visible fiber structures, and fragmentation reducing the accuracy of
traditional binarization methods for distinguishing text from
background. DL approaches are able to effectively binarize degraded
images because they learn context-dependent features. Traditional
methods, in contrast, depend on histograms or local statistics, making
them more susceptible to errors when dealing with image degradations
\citep{christleinBinCompare}.

The central aim of this work is to systematically evaluate the impact
of binarization techniques on the performance of writer identification,
particularly in the context of Greek papyrus.  Also, an augmentation is proposed to enhance binarization. 
To summarize, the key contributions are:

\begin{itemize}
\item
  A data augmentation methodology to adapt existing binarization
  datasets for improved performance on Greek papyri.
\item
  An analysis between binarization quality and subsequent
  writer identification performance, demonstrating that the PSNR metric
  on papyri (DIBCO 2019 \citep{dibco2019} Set B) serves as the most
  effective indicator for selecting binarization models that optimize
  writer identification.
\item
  The identification of the most effective model selection metric (F-Measure)
  for training DL-based binarization models in the context of papyri.
\end{itemize}

The paper is organized as follows. Section \ref{methodology} describes the
datasets and presents the methodology for the proposed augmentation technique, the training of DL binarization models, and the subsequent writer identification approach. Section \ref{evaluation_and_results} presents and discusses the results for both binarization and writer identification, as well as an analysis of their correlation, and Section \ref{conclusion} concludes the paper.
\section{Methodology} \label{methodology}
This section details the methodology employed for training and
evaluating DL models binarization performance. It outlines the datasets
used for training and validation, as well as the developed augmentation
method. Subsequently, the writer identification pipeline, which is based
on the approach proposed by Christlein et al.
\citep{christlein2022writer}, is described. 

\subsection{Datasets}

The dataset used for training and validation of the DL binarization
models includes the following datasets: \textbf{DIBCO} (2009-2013, 2017), \textbf{H-DIBCO} (2014, 2016, 2018) \citep{6981120}\citep{7814134}\citep{dibco2018}, \textbf{BICKLEY} a collection of Ms. Anna Felton   Bickley's handwritten diary (1922 to 1926) \citep{deng2010binarizationshop}, \textbf{Monk Cuper Set} (MCS) containing a series of 17th-century scholarly handwritten letters (in multiple languages) \citep{deepotsu}, \textbf{Handwritten Malayalam palm leaf manuscript dataset} (HMPLMD) containing images of palm leaves inscribed with Malayalam script \citep{HMPLMD}, and \textbf{Persian Heritage Image Binarization Dataset} (PHIBD) containing images of historical Persian manuscripts from a library in Yazd, Iran,  which also exhibit various types of degradation, including bleed-through, faded ink and blur \citep{phibd}.

The DIBCO 2019 \citep{dibco2019} dataset used for DL model binarization
performance evaluation consists of 20 image-ground truth pairs,
partitioned into two parts: Set A and Set B, each containing 10
image-ground truth pairs. Set A includes images representing common
challenges found within the DIBCO series, containing images featuring
predominantly black text on white background, while Set B is exclusively
composed of images of papyri fragments. 

For writer retrieval and identification evaluation, the GRK-Papyri
dataset \citep{grk_papyri} is utilized. This dataset contains 50 labeled
images of Greek papyri fragments from approximately 600 CE, where each
image is attributed to one out of the 10 scribes.

\subsection{Data Augmentation}

A data augmentation method is developed to enhance DL models' performance
on papyri. The objective is to specifically adapt the training dataset
by simulating key visual characteristics found in papyrus documents such
as fiber structure. This involves applying two primary augmentations to
existing training images: Firstly, the application of papyrus textures such as those found in
collections like GRK-Papyri \citep{grk_papyri} and secondly, the simulation of the irregular, degraded edges characteristic of
papyrus fragments.

When augmenting the dataset, HMPLMD \citep{HMPLMD} is excluded, since its
images are already similar in color and texture. For the augmentation
GRK-120 \citep{grk_papyri}
is used, which is an extension of the GRK-Papyri \citep{grk_papyri}
dataset.

\begin{figure}
\centering
\includegraphics[width=0.7\textwidth]{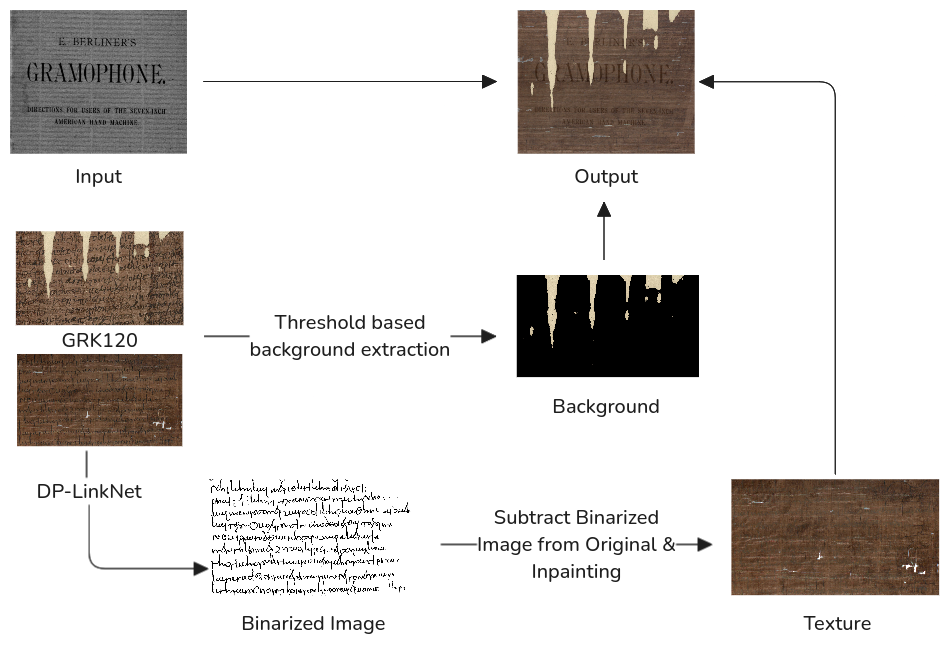}
\caption{Two Step Augmentation Process}
\end{figure}

The first augmentation involves recreating a similar background
featuring fiber structure by removing writing from images sampled from
GRK-120 \citep{grk_papyri}. A pretrained DP-LinkNet \citep{dplinknet} model
 generates a binary mask identifying the
text. The TELEA inpainting algorithm
\citep{Telea01012004}\citep{opencv_library} is then applied, using this
mask and a 5-pixel radius, to reconstruct the background.

The second augmentation aims to replicate the non-uniform edges
characteristic of papyrus fragments. For this step, images are again
sampled from GRK-120 \citep{grk_papyri} and the background surrounding
the papyrus is extracted. An empirically determined brightness threshold
of 170 (on the image's 8-bit grayscale representation, 0-255) is used to
identify background pixels, as the majority of GRK-120 \citep{grk_papyri}
contains bright backgrounds. Images with background brightness below
this threshold are not used. 


The final augmented image is created by first resizing the background papyrus texture from the initial augmentation step and converting it to RGBA format. Its transparency (alpha channel) is set to 70\%, a value chosen empirically to balance texture visibility and preservation of the original image details. This semi-transparent texture is then alpha-composited onto the original image and converted back to RGB format. Lastly, the RGBA background image from the second augmentation step is overlaid onto this textured image to simulate the fragment's shape. This approach leverages existing handwriting datasets while facilitating training tailored to specific tasks.

\subsection{Binarization Model Training and Evaluation}


Each DL binarization model is trained using the hyperparameters
recommended in its respective original publication. Model selection
refers to the criteria used to determine which models are saved during
training. By monitoring F-Measure (FM), pseudo-F-Measure (pFM), and
Peak-Signal-To-Noise-Ratio (PSNR) scores achieved on the validation set,
the best performing model per metric is saved. These metrics are
monitored for all networks except DeepOtsu \citep{deepotsu}. For the
evaluation, metrics for binarization performance are adopted from DIBCO
2019 \citep{dibco2019}. 
For comparative analysis, traditional binarization approaches are
implemented using default parameters from the doxapy library\footnote{\url{https://github.com/brandonmpetty/Doxa}}.

\subsection{Writer Identification Pipeline}

The unsupervised feature learning approach proposed by Christlein et al.
\citep{christlein2022writer} has been used to evaluate the impact of binarization on writer retrieval and writer classification. 
Patches of size 32 x 32 are extracted at R-SIFT \citep{christlein2022writer} keypoint positions ( >95\% black pixels) and paritioned in 5000 clusters. Feature extraction is done by a ResNet-20 \citep{resnet34} which is trained for each binarization method and Vectors of Locally Aggregated Descriptors (VLAD) are used to create a global feature vector (128 clusters).

For writer retrieval, cosine similarity between VLAD representations is
measured. Each query image's VLAD representation is then compared against the representations of the remaining 49 images in the dataset. For writer classification, two images per writer serve as the reference
set for a Nearest-Neighbor (NN) classifier. Performance is then
evaluated by classifying the remaining images based on their similarity
to these reference samples. However, this approach carries the risk of
producing unrepresentative results, as different image selections
perform significantly better or worse than others. To mitigate this
issue, a set of 500 unique random combinations of training images is
precomputed and applied for each method, and the results are then
averaged to ensure a more representative evaluation of performance.


\section{Evaluation and Results} \label{evaluation_and_results}
This section assesses the performance of binarization approaches and
their effects on downstream writer retrieval and identification tasks.

\subsection{Binarization}

This section contains the performance evaluation of binarization
approaches applied to the DIBCO 2019 \citep{dibco2019} dataset. Results
are presented for Set B, comparing the effectiveness of DL approaches
against traditional methods. The analysis further investigates the
impact of data augmentation on DL model performance, the influence of
different model selection criteria, and the effect of parameter
optimization on traditional binarization methods.

\begin{table}[h!]
    \centering
    \caption{Results for Best Performing Models (Impact of Augmentation shown in brackets)}
    \label{tab:augmentation_binarization}
    \begin{tabular}{l l l l l}
        \toprule
        Approach & FM $\uparrow$ & pFM $\uparrow$ & PSNR $\uparrow$ & DRD $\downarrow$ \\
        \midrule
        DP-LinkNet \citep{dplinknet} & 79.8 (+19.0) & 79.1 (+18.0) & 15.1 (+1.3) & 10.3 (-5.6) \\
        DeepOtsu \citep{deepotsu} & 72.0 (+1.0) & 69.8 (-0.4) & 13.6 (+0.0) & 19.5 (+2.1) \\
        NAF-DPM \citep{nafdpm} & 69.3 (+14.6) & 69.3 (+11.0) & 14.1 (+1.4) & 15.9 (-3.3) \\
        Robin \citep{robin} & 67.3 (+10.4) & 67.4 (+9.9) & 14.8 (+2.1) & 13.2 (-6.5) \\
        \bottomrule
    \end{tabular}
\end{table}

Augmentation improves performance of all DL approaches significantly,
except DeepOtsu \citep{deepotsu}, as shown in Table~\ref{tab:augmentation_binarization}. While DP-LinkNet \citep{dplinknet} gains
up to 19 percentage points in FM, DeepOtsu \citep{deepotsu} gains only
1.0 in FM but performance worsens slightly in pFM and DRD. The marginal
gain and partial decline in DeepOtsu \citep{deepotsu} performance may stem
from its unique architecture, as it is the only model evaluated that
does not use an end-to-end encoder-decoder structure for binarization.
Instead, it initially focuses on enhancing the background by removing
degradation, an approach that may be less effective considering the
degradations specific to papyri.


To investigate whether the lower performance of traditional methods on
Set B of DIBCO 2019 \citep{dibco2019} results from suboptimal parameter
settings, a parameter search is conducted with window sizes ranging from
37 to 147, minN values spanning the same range, and glyph parameters
varying from 30 to 120. This search presented in Table~\ref{tab:traditional_parameter_tuning} compares the results of the traditional algorithms obtained on
DIBCO 2019 \citep{dibco2019} Set B using the optimized parameters.


\begin{table}[h!]
\centering
\caption{Traditional Binarization Methods using Default versus Optimized Parameters}
\label{tab:traditional_parameter_tuning}
\begin{tabular}{l c c | c c c c}
\toprule
Approach & Window Size & Additional & \textit{FM $\uparrow$} & \textit{pFM $\uparrow$} & \textit{PSNR $\uparrow$} & \textit{DRD $\downarrow$} \\
         &             & Parameter  &                         &                         &                            &                           \\
\midrule
Gatos \citep{GATOS2006317} & 37 & Glyph: 110 & 62.8 & 63.7 & 11.7 & 26.6 \\
Nick \citep{nick} & 47 & - & 61.5 & 62.7 & 11.5 & 28.0 \\
Trsingh \citep{trsingh} & 47 & - & 60.8 & 61.7 & 11.3 & 29.4 \\
Sauvola \citep{sauvola} & 37 & - & 57.3 & 57.5 & 10.6 & 36.1 \\
Su \citep{su} & 37 & minN: 147 & 56.8 & 57.3 & 10.2 & 48.0 \\
Otsu \citep{otsu} & - & - & 23.3 & 20.7 & 2.8 & 209.3 \\
\bottomrule
\end{tabular}
\end{table}

\begin{figure}[h!]
    \centering
    \begin{subfigure}[b]{0.3\textwidth}
        \centering
        \includegraphics[width=\textwidth]{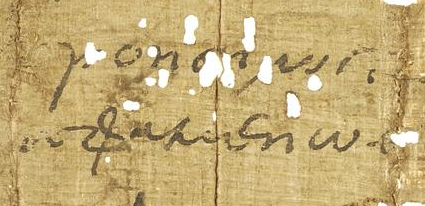}
        \caption{Original (Menas\_1)}
        \label{fig:original}
    \end{subfigure}
    \hfill
    \begin{subfigure}[b]{0.3\textwidth}
        \centering
        \includegraphics[width=\textwidth]{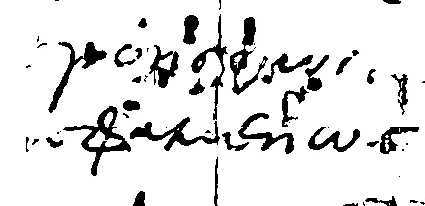}
        \caption{Robin}
        \label{fig:robin-fmeasure}
    \end{subfigure}
    \hfill
    \begin{subfigure}[b]{0.3\textwidth}
        \centering
        \includegraphics[width=\textwidth]{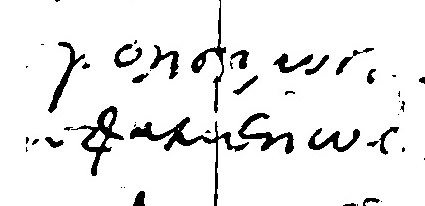}
        \caption{Robin + Augm.}
        \label{fig:enhanced-robin}
    \end{subfigure}
    
    \vspace{2mm}
    
    \begin{subfigure}[b]{0.3\textwidth}
        \centering
        \includegraphics[width=\textwidth]{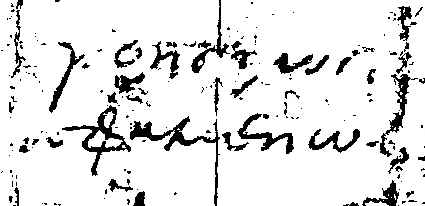}
        \caption{Gatos}
        \label{fig:gatos}
    \end{subfigure}
    \hfill
    \begin{subfigure}[b]{0.3\textwidth}
        \centering
        \includegraphics[width=\textwidth]{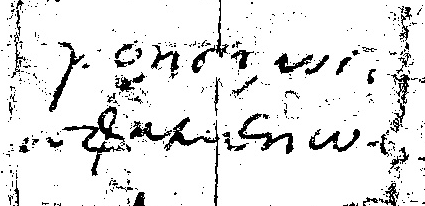}
        \caption{Nick}
        \label{fig:nick}
    \end{subfigure}
    \hfill
    \begin{subfigure}[b]{0.3\textwidth}
        \centering
        \includegraphics[width=\textwidth]{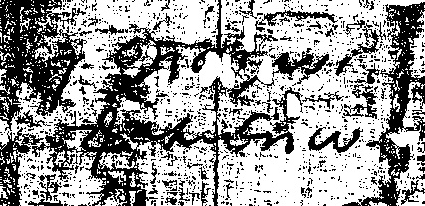}
        \caption{Otsu}
        \label{fig:otsu}
    \end{subfigure}
    
    \vspace{2mm}
    
    \begin{subfigure}[b]{0.3\textwidth}
        \centering
        \includegraphics[width=\textwidth]{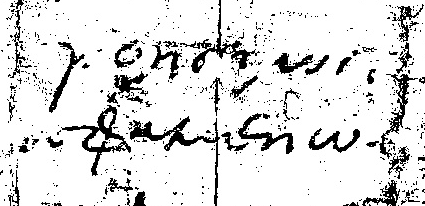}
        \caption{Sauvola}
        \label{fig:sauvola}
    \end{subfigure}
    \hfill
    \begin{subfigure}[b]{0.3\textwidth}
        \centering
        \includegraphics[width=\textwidth]{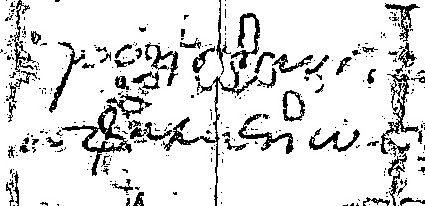}
        \caption{Su}
        \label{fig:su}
    \end{subfigure}
    \hfill
    \begin{subfigure}[b]{0.3\textwidth}
        \centering
        \includegraphics[width=\textwidth]{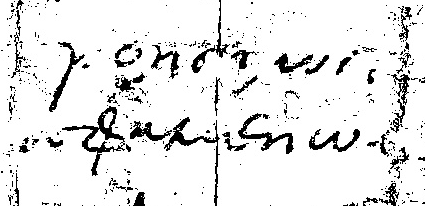}
        \caption{Trsingh}
        \label{fig:trsingh}
    \end{subfigure}
    
    \vspace{2mm}
    
    \begin{subfigure}[b]{0.3\textwidth}
        \centering
        \includegraphics[width=\textwidth]{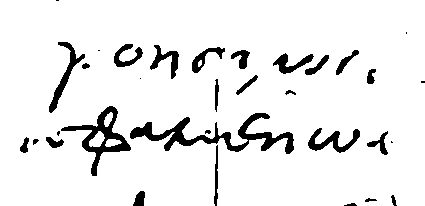}
        \caption{DP-LinkNet + Augm.}
        \label{fig:enhanced-dplinknet}
    \end{subfigure}
    \hfill
    \begin{subfigure}[b]{0.3\textwidth}
        \centering
        \includegraphics[width=\textwidth]{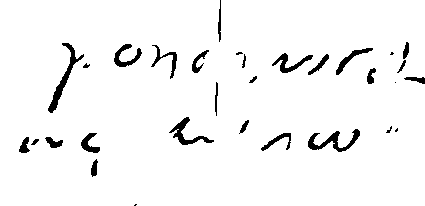}
        \caption{NAF-DPM + Augm.}
        \label{fig:enhanced-nafdpm}
    \end{subfigure}
    \hfill
    \begin{subfigure}[b]{0.3\textwidth}
        \centering
        \includegraphics[width=\textwidth]{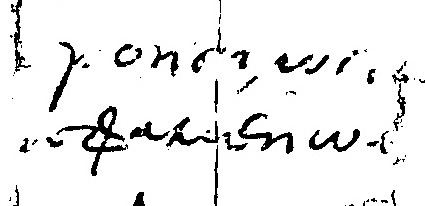}
        \caption{DeepOtsu + Augm.}
        \label{fig:enhanced-unetbr}
    \end{subfigure}
        \caption{Illustration of different binarization approaches and the impact of augmentation.}    
    \label{fig:binarization_visual_augmented_full}
\end{figure}

Across the
methods tested, the parameter search suggests that a window size between
37 and 47 is optimal, although potential improvements from lower,
untested window sizes cannot be ruled out. Notably, character sizes in
DIBCO 2019 \citep{dibco2019} vary significantly. In Set A, the majority of
characters ranged between 10 and 20 pixels in height, whereas in Set B,
approximately 50\% of characters measured 75 pixels or taller, with the
remainder ranging from 20-50 pixels. However, even with these
improvements, the optimized traditional methods still underperform
compared to the leading Deep Learning models. This performance gap is
particularly evident in their notably higher DRD scores. A qualitative example of the binarization of all approaches evaluated is given in Figure~\ref{fig:binarization_visual_augmented_full}.

\subsection{Writer Identification}

This section evaluates the impact of the previously evaluated binarization methods on the
downstream tasks of writer retrieval and classification. First,
traditional binarization methods are examined, and the top approaches are compared to SOTA methods. Following this,
the impact of different model selections as well as the impact of
augmentation on writer identification performance is evaluated.

Performance is assessed using the GRK-Papyri \citep{grk_papyri} dataset,
employing mean Average Precision (mAP) and Top-1 accuracy for writer
retrieval, and Top-1 and Top-5 accuracy for writer classification.

\subsubsection{Evaluation of Traditional Binarization Methods}

\begin{table}[h!]
    \centering
    \caption{Traditional Identification and Retrieval Results}
    \label{tab:traditional_id_retrieval}
    \begin{tabular}{l c c | c c}
        \toprule
        \multirow{2}{*}{Approach} & \multicolumn{2}{c}{Writer Classification} & \multicolumn{2}{c}{Writer Retrieval} \\
        \cmidrule(lr){2-3} \cmidrule(lr){4-5} 
        & \textit{Top-1} & \textit{Top-5} & \textit{mAP} & \textit{Top-1} \\
        \midrule 
        Trsingh \citep{trsingh} & 34.2 & 72.7 & 33.0 & 38.0 \\
        Gatos \citep{GATOS2006317} & 33.2 & 69.4 & 32.1 & 38.0 \\
        Nick \citep{nick} & 32.3 & 68.0 & 30.7 & 38.0 \\
        Sauvola \citep{sauvola} & 32.0 & 70.9 & 31.4 & 36.0 \\
        Su \citep{su} & 23.3 & 59.3 & 25.4 & 30.0 \\
        Otsu \citep{otsu} & 22.5 & 61.6 & 24.1 & 28.0 \\
        \bottomrule
    \end{tabular}
\end{table}

A notable performance gap is evident in Table \ref{tab:traditional_id_retrieval}: Trsingh
\citep{trsingh}, Gatos \citep{GATOS2006317}, Nick \citep{nick}, and Sauvola
\citep{sauvola} consistently outperform Su \citep{su} and Otsu \citep{otsu}
across all evaluated metrics. Specifically, for writer retrieval, they
achieve mAP scores above 30\%, with Trsingh leading at 33.03\%. In
contrast, Su and Otsu yield significantly lower mAP scores of 25.35\%
and 24.10\%, respectively. This trend holds for writer classification,
where Trsingh again achieves the highest Top-1 accuracy of 34.17\%,
outperforming Su \citep{su} with 23.33\% and Otsu \citep{otsu} with 22.47\%.
These lower scores reflect inherent limitations: Otsu's \citep{otsu}
global thresholding struggles with non-uniform backgrounds, while the
effectiveness of Su's \citep{su} contrast-based method may be compromised
by severe degradations found in papyri.

\subsubsection{Comparison to SOTA Writer Identification Approaches}

\begin{table}[h!]
    \centering
    \caption{Comparison to SOTA Writer Classification Approaches}
    \label{tab:sota_retrieval_comparison}
    \begin{tabular}{l l c c c c c}
        \toprule
        Approach & Model           & Augmentation & Top-1 & STD & Top-5 & STD \\
                 & Selection       &              &       &     &       &     \\
        \midrule
        Robin \citep{robin} & FM & \checkmark & 49.0 & 7.9 & 80.0 & 7.5 \\
        NAF-DPM \citep{nafdpm} & FM & \checkmark & 46.5 & 7.4 & 77.3 & 7.3 \\
        DP-LinkNet \citep{dplinknet} & FM & \checkmark & 46.3 & 8.3 & 77.5 & 7.6 \\
        DP-LinkNet \citep{dplinknet} & PSNR & \checkmark & 45.2 & 7.4 & 75.0 & 7.6 \\
        Robin \citep{robin} & pFM & \checkmark & 45.2 & 8.4 & 79.4 & 7.4 \\
        \midrule
        AngU-Net + R-SIFT + NN \citep{christlein2022writer} &  &  & 53.0 & - & 77.0 & - \\
        AngU-Net + SIFT + NN \citep{christlein2022writer} &  &  & 47.0 & - & 83.0 &  \\ 
        \bottomrule
    \end{tabular}
\end{table}

Robin \citep{robin} utilizing FM model selection achieves a Top-1 accuracy
4 percentage points lower than the architecturally similar AngUNet +
R-SIFT + NN method reported by Christlein et al.
\citep{christlein2022writer}. However, performance variability is high,
with standard deviations ranging from 7 to 9 percent, as observed across
500 tested random combinations of training images for the NN classifier.
This variability likely results from the sensitivity of the classifier
to specific training subsets, compounded by the limited dataset size of
GRK-Papyri \citep{grk_papyri}, which contains only 50 images from 10
scribes.


\begin{table}[h!]
    \centering
    \caption{Comparison to SOTA Writer Retrieval Approaches}
    \label{tab:retrieval_results}
    \begin{tabular}{l c c c c c c}
        \toprule
        Approach & Model Selection & Augmentation & Top-1 & Top-5 & Top-10 & mAP \\
        \midrule
        Robin \citep{robin} & FM & \checkmark & 56.0 & 86.0 & 94.0 & 42.4 \\
        Robin \citep{robin} & pFM & \checkmark & 54.0 & 78.0 & 90.0 & 41.3 \\
        DP-LinkNet \citep{dplinknet} & FM & \checkmark & 52.0 & 80.0 & 86.0 & 40.9 \\
        NAF-DPM \citep{nafdpm} & FM & \checkmark & 52.0 & 86.0 & 88.0 & 40.8 \\
        Robin \citep{robin} & PSNR & \checkmark & 48.0 & 78.0 & 84.0 & 40.5 \\
        \midrule
        AngU-Net + Cl-S \citep{8270096} &  &  & 52.0 & 82.0 & 94.0 & 42.2 \\
        AngU-Net + R-SIFT \citep{christlein2022writer} &  &  & 48.0 & 84.0 & 92.0 & 42.8 \\
        \bottomrule
    \end{tabular}
\end{table}

When evaluated against the AngU-Net + R-SIFT method, the Robin
\citep{robin} approach utilizing FM model selection and augmentation
emerges again as the most effective. Although its mAP is 0.4 percentage points lower
compared to AngU-Net + R-SIFT, Robin \citep{robin} achieves
state-of-the-art (SOTA) performance in Top-1 accuracy, reaching 56\%.
Robin \citep{robin} configurations constitute 2 of the top 5 approaches
for writer identification (See Table \ref{tab:sota_retrieval_comparison}) and 3 of the top 5 for writer retrieval (See Table \ref{tab:retrieval_results}). Considering this, Robin \citep{robin} emerges as the best performing model for writer identification among the
evaluated binarization approaches.

\subsubsection{Evaluation of Model Selection}
\begin{table}[h!]
    \centering
        \caption{Model Selection Results}
    \label{tab:model_selection_combined}
    \begin{subtable}{0.48\textwidth}
        \centering
        \caption{Classification}
        \label{tab:model_selection_classification}
        \begin{tabular}{l c c}
            \toprule
            Model Selection & Top-1 & Top-5 \\
            \midrule
            FM    & 47.3 & 78.3 \\
            PSNR  & 42.9 & 76.2 \\
            pFM   & 41.6 & 77.9 \\
            \bottomrule
        \end{tabular}
    \end{subtable}
    \hfill
    \begin{subtable}{0.48\textwidth}
        \centering
        \caption{Retrieval}
        \label{tab:model_selection_retrieval}
        \begin{tabular}{l c c}
            \toprule
            Model Selection & mAP & Top-1 \\
            \midrule
            FM    & 41.3 & 53.3 \\
            pFM   & 39.0 & 47.3 \\
            PSNR  & 38.7 & 46.7 \\
            \bottomrule
        \end{tabular}
    \end{subtable}

\end{table}

Table \ref{tab:model_selection_combined} details the impact of different model selection criteria
during DL binarization training on subsequent writer retrieval
performance. A clear performance advantage is observed for FM, which
achieves a mAP score of 41.3. In contrast, using either pFM or PSNR as
the selection criterion resulted in comparatively lower performance,
with mAP scores slightly below 39\% for both. This advantage for FM is
further reinforced by its significantly higher Top-1 accuracy compared
to the other criteria. Therefore, FM emerges as the most effective model
selection criterion when training DL binarization approaches
specifically to optimize performance for subsequent writer retrieval
tasks.

For writer classification tasks, FM again demonstrates superior
performance (See Table \ref{tab:model_selection_classification}), confirming its effectiveness as a model selection criterion
also for writer classification. In contrast to the results observed in Table
\ref{tab:model_selection_retrieval}, the relative ranking of PSNR and pFM changes in this
context, with PSNR achieving slightly higher Top-1 scores than pFM.
However, both PSNR and pFM yield results noticeably lower than those
obtained using FM.

\subsubsection{Evaluation of Augmentation}

\begin{table}[h!]
    \centering
    \caption{Impact of Augmentation}
    \begin{subtable}{0.4\textwidth}
        \centering
        \caption{Classification}
        \label{tab:augmentation_writer_classification}
        \begin{tabular}{l l}
            \toprule
            Approach & Top-1 \\
            \midrule
            Robin \citep{robin}       & 49.0 (+14.4) \\
            NAF-DPM \citep{nafdpm}    & 46.5 (+13.9) \\
            DP-LinkNet \citep{dplinknet} & 46.3 (+8.2) \\
            DeepOtsu \citep{deepotsu} & 34.0 (-2.7) \\
            \bottomrule
        \end{tabular}
    \end{subtable}
    \hfill
    \begin{subtable}{0.55\textwidth}
        \centering
        \caption{Retrieval}
        \label{tab:augmentation_writer_retrieval}
        \begin{tabular}{l ll}
            \toprule
            Approach & mAP & Top-1 \\
            \midrule
            Robin \citep{robin}       & 42.4 (+8.8)  & 56.0 (+22.0) \\
            DP-LinkNet \citep{dplinknet} & 40.9 (+5.2)  & 52.0 (+12.0) \\
            NAF-DPM \citep{nafdpm}    & 40.8 (+8.9)  & 52.0 (+20.0) \\
            DeepOtsu \citep{deepotsu} & 32.1 (-1.9)  & 34.0 (-8.0) \\
            \bottomrule
        \end{tabular}
    \end{subtable}

    \label{tab:augmentation_writer_combined}
\end{table}

In writer classification, DL models surpass traditional methods even
without augmentation (See Tables \ref{tab:traditional_id_retrieval},
\ref{tab:augmentation_writer_classification}). Data augmentation
significantly widens this performance gap, with DL models achieving
Top-1 accuracy scores of up to 49.03\%. Consistent with previous
observations, augmentation benefits all tested DL models apart from
DeepOtsu \citep{deepotsu}.

Without data augmentation, the performance of DL approaches is
comparable to traditional methods (See Tables \ref{tab:traditional_id_retrieval},
\ref{tab:augmentation_writer_retrieval}). However, augmentation
significantly improved results for most DL models. Robin \citep{robin} and
NAF-DPM \citep{nafdpm} increased their mAP by approximately 9 percentage
points with augmentation, although DeepOtsu's \citep{deepotsu} mAP
decreased by 1.86 percentage points.

\subsection{Correlation}

\begin{figure}[h!]
    \centering
    \caption{Scatter Plots Across DIBCO 2019 Subsets for Correlations Between Binarization Metrics and Writer Retrieval mAP}
    \begin{subfigure}[b]{0.45\textwidth}
        \centering
        \includegraphics[width=\textwidth]{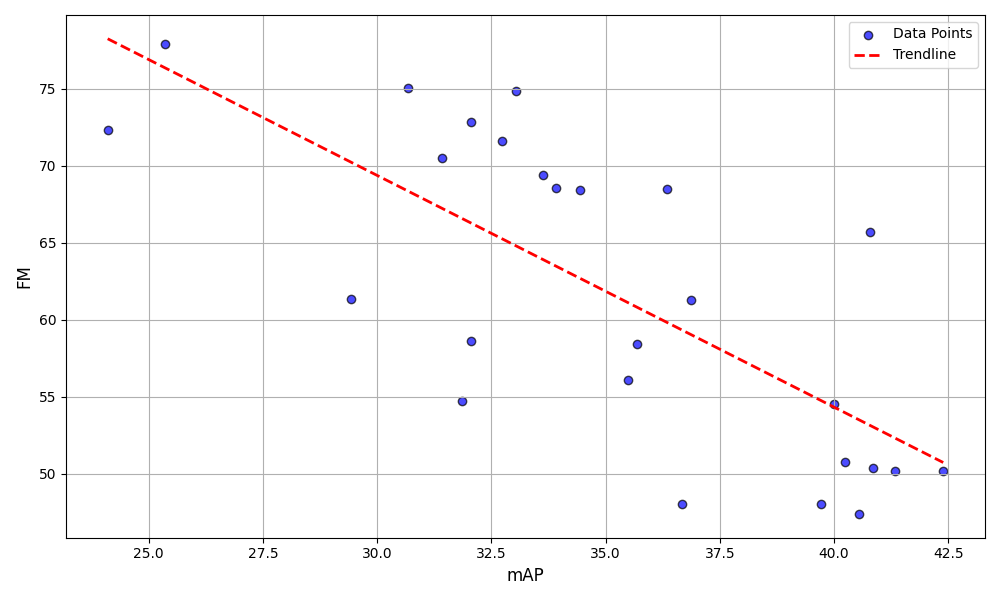}
        \caption{Correlation between FM on DIBCO 2019 Set A and writer retrieval mAP (r = -0.734, p < 0.001).}
        \label{fig:fm-a}
    \end{subfigure}
    \hfill
    \begin{subfigure}[b]{0.45\textwidth}
        \centering
        \includegraphics[width=\textwidth]{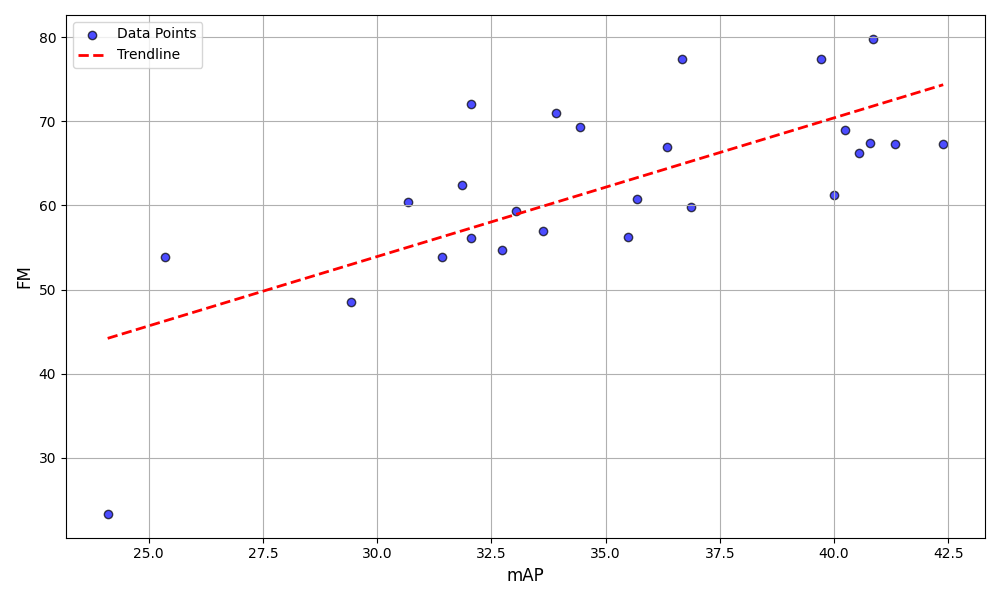}
        \caption{Correlation between FM on DIBCO 2019 Set B and writer retrieval mAP (r = 0.711, p < 0.001).}
        \label{fig:fm-b}
    \end{subfigure}
    
    \vspace{1em} 
    
    \begin{subfigure}[b]{0.45\textwidth}
        \centering
        \includegraphics[width=\textwidth]{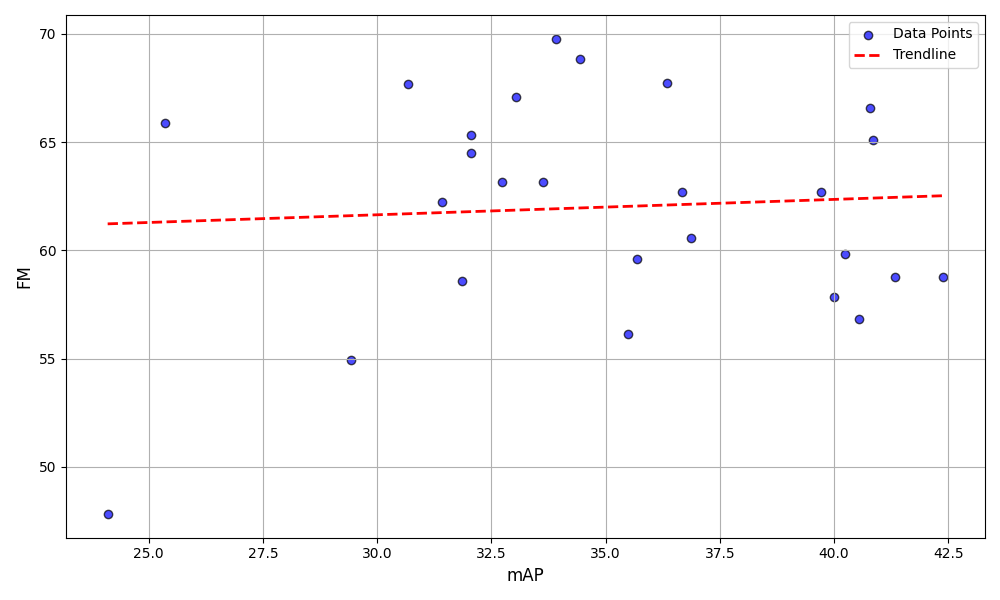}
        \caption{Negligible correlation between FM on DIBCO 2019 Set A+B and writer retrieval mAP (r = 0.068, p = 0.74).}
        \label{fig:fm-a-b}
    \end{subfigure}
    \hfill
    \begin{subfigure}[b]{0.45\textwidth}
        \centering
        \includegraphics[width=\textwidth]{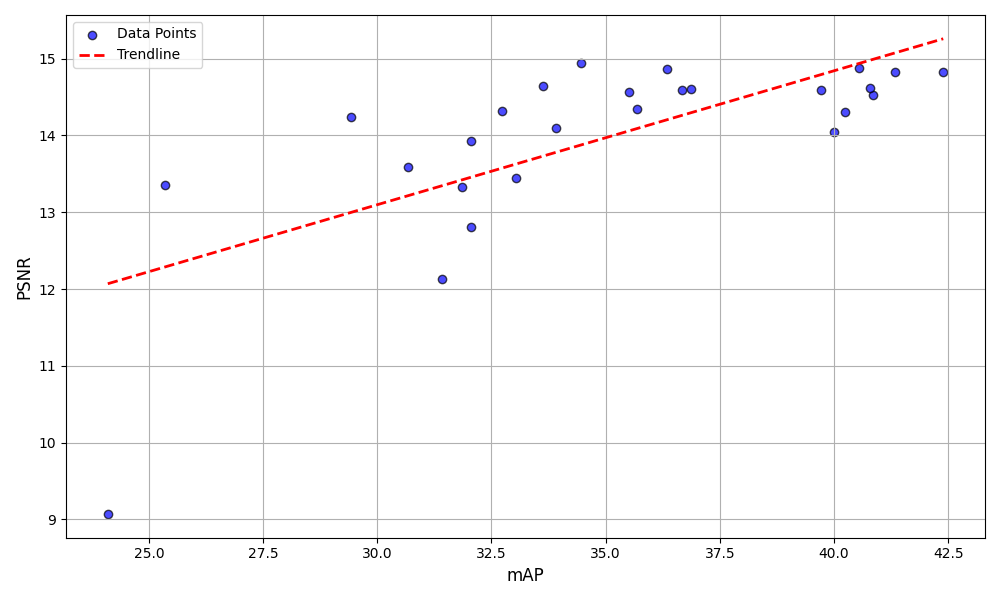}
        \caption{Strong positive correlation between PSNR on DIBCO 2019 Set A+B and writer retrieval mAP (r = 0.695, p < 0.001).}
        \label{fig:psnr-a-b}
    \end{subfigure}
    
    \label{fig:correlation-plots}
\end{figure}

To investigate the relationship between binarization performance and the
performance of subsequent writer analysis tasks, a correlation analysis
is performed. This involves correlating DIBCO 2019 \citep{dibco2019}
binarization performance metrics with subsequent writer retrieval (mAP) and
identification performance (Top-1). 
Figure \ref{fig:correlation-plots} shows the details of the correlation analysis. It indicates consistent patterns across both writer identification and retrieval tasks. For Set A, FM and pFM metrics exhibit strong, significant negative correlations ($r = -0.73$) with writer analysis performance, while PSNR has a moderate negative correlation ($r \approx -0.46$). Conversely, Set B shows strong positive correlations for PSNR ($r \approx 0.78$) and FM/pFM ($r \approx 0.72$), with DRD negatively correlated ($r \approx -0.60$). Combining Sets A and B neutralizes the significance of FM/pFM correlations due to opposing dataset-specific trends, but PSNR remains consistently and strongly positively correlated ($r \approx 0.68$), highlighting PSNR as a robust indicator of performance.

\section{Conclusion}
This paper examined how image binarization quality affects writer identification performance, showing that DL-based methods outperform traditional approaches. A strong correlation was found between PSNR scores on Set B of DIBCO 2019 and writer identification outcomes, resulting in PSNR as a reliable predictor. FM emerged as the most effective criterion for writer classification and retrieval.

Data augmentation generally enhanced writer identification performance, with the most notable gains seen in Robin and NAF-DPM. However, results are based on a small dataset, which may limit generalizability. Future work should focus on larger datasets and more advanced augmentation tailored to Greek papyri.

 \label{conclusion}

\medskip
{
\small
\bibliography{sources}
}


\end{document}